\documentclass[sigconf, nonacm, balance=false]{acmart}
\AtBeginDocument{%
  }

\setcopyright{none}
\acmDOI{}
\acmConference[Under-review]{Under-review}{}{}
\acmISBN{}

\newcommand{\arxivpageformat}{%
  \fancyhf{}%
  \fancyhead[C]{\sffamily\footnotesize Under-review}%
  \fancyfoot[C]{\sffamily\footnotesize\thepage}%
  \renewcommand{\headrulewidth}{0pt}%
  \renewcommand{\footrulewidth}{0pt}%
}
\AtBeginDocument{%
  \fancypagestyle{standardpagestyle}{\arxivpageformat}%
  \fancypagestyle{firstpagestyle}{\arxivpageformat}%
  \pagestyle{standardpagestyle}%
}

\usepackage{graphicx}
\usepackage{booktabs}

\usepackage{url}
\usepackage{array}
\usepackage{multirow}
\usepackage{enumitem}
\usepackage{amsmath}
\usepackage{csquotes}  
\def\onedot{.\xspace}
\usepackage{xspace}
\definecolor{orangeyellow}{RGB}{255, 204, 0}
\definecolor{cvprblue}{rgb}{0.21,0.49,0.74}
\definecolor{nvidiagreen}{RGB}{119,185,0}
\usepackage{pifont}  
\usepackage{tabularx}
\usepackage{float}
\usepackage[normalem]{ulem}

\begin{document}
\def\etal{\emph{et~al.}}
\def\eg{\emph{e.g}\onedot} \def\Eg{\emph{E.g}\onedot}
\def\ie{\emph{i.e}\onedot} \def\Ie{\emph{I.e}\onedot}
\def\cf{\emph{c.f}\onedot} \def\Cf{\emph{C.f}\onedot}
\def\etc{\emph{etc}\onedot} \def\vs{\emph{vs}\onedot}
\def\wrt{w.r.t\onedot} \def\dof{d.o.f\onedot}

\newcommand{\zznote}[1]{\textcolor{magenta}{ZZ:#1}}
\newcommand{\zzdel}[1]{\textcolor{red}{\sout{#1}}}
\newcommand{\zzrep}[2]{\textcolor{red}{\sout{#1}}\textcolor{blue}{#2}}
\newcommand{\zzadd}[1]{\textcolor{blue}{#1}}
\newcommand\hunote[1]{\textcolor{orangeyellow}{hu: #1}}
\newcommand\haonote[1]{\textcolor{blue}{HAO: #1}}
\newcommand{\reffig}[1]{Figure~\ref{#1}}
\newcommand{\reftab}[1]{Table~\ref{#1}}
\newcommand{\refsection}[1]{Section~\ref{#1}}
\newcommand{\refeq}[1]{Eq.~(\ref{#1})}
\newlength\savewidth\newcommand\shline{\noalign{\global\savewidth\arrayrulewidth
\global\arrayrulewidth 1pt}\hline\noalign{\global\arrayrulewidth\savewidth}}
\newcommand\todohao[1]{\colorbox{yellow}{#1}}
\definecolor{verbColor}{HTML}{F29E6E}
\definecolor{colorColor}{HTML}{C7A446}
\title{AnomalyLMM: Bridging Generative Knowledge and Discriminative Retrieval for Text-based Person Anomaly Search}

\author[Hao Ju et al.]{\texorpdfstring{%
  Hao Ju\textsuperscript{1}\quad
  Hu Zhang\textsuperscript{2}\quad
  Zhedong Zheng\textsuperscript{1}%
}{Hao Ju; Hu Zhang; Zhedong Zheng}}
\affiliation[obeypunctuation=true]{%
  \institution{\textsuperscript{1}University of Macau\\
    \textsuperscript{2}CSIRO Data61}%
  \city{}%
  \country{}%
}

\renewcommand{\shortauthors}{Hao Ju, Hu Zhang, and Zhedong Zheng}

\begin{abstract}
Text-based person anomaly search requires fine-grained alignment between language queries and subtle visual cues, a challenge amplified by the long-tailed nature of anomalous behaviors. While Large Multimodal Models (LMMs) offer powerful visual understanding, their generative pre-training is fundamentally misaligned with the discriminative needs of this retrieval task. Adapting them without costly fine-tuning remains a significant hurdle. We introduce \textbf{AnomalyLMM}, a training-free, coarse-to-fine framework that effectively repurposes generative LMMs for zero-shot anomaly retrieval. Our method first uses a general retrieval model to produce an initial ranking. To refine this list, we introduce a novel cloze-based re-ranking mechanism with three steps. The first step \textbf{Cloze Generation} converts the text query into a ``fill-in-the-blank'' prompt. Next, \textbf{Cloze Completion} compels the LMM to focus on specific visual regions and generate a description of the potential anomaly. The final step \textbf{Comparison \& Re-ranking} calculates semantic alignment between the LMM's generated completion and the original query, which serves as a powerful re-ranking score.
Experiments on the PAB and UCC datasets show that AnomalyLMM surpasses the competitive baseline by $+1.93$\% and $+1.79$\% Recall@1 accuracy on two datasets. Crucially, our method provides highly interpretable visual-textual alignments without any task-specific training, a vital feature for real-world deployment. The code will be made publicly available.
\end{abstract}




\maketitle
\noindent\textbf{Web Relevance:}
This paper focuses on improving fine-grained text-to-image retrieval for Web-based visual content, enabling more accurate and interpretable multimedia search through large multimodal models.
\section{Introduction}
Ensuring public safety is a growing priority in smart surveillance and web systems, driving the need for efficient text-based person anomaly search, a task that retrieves individuals exhibiting abnormal behaviors using natural language queries. Unlike generic text-based person search~\cite{zheng2020dual,yang2023towards}, this task focuses on identifying subtle, often rare anomalous actions (\eg, falling, collision, or being hit) described in text. Such capability is critical for real-world applications, from crime prevention to emergency response. However, it remains underexplored due to the lack of dedicated benchmarks and the inherent complexity of aligning textual descriptions with fine-grained anomalous behaviors in videos.

\begin{figure*}[t]
    \centering
    \includegraphics[width=0.98\textwidth]{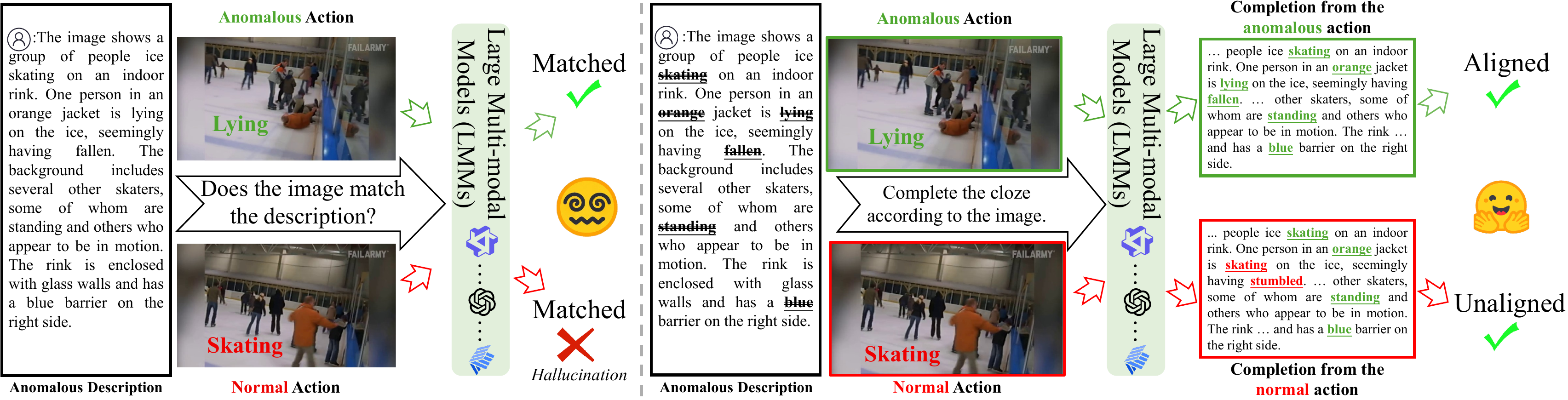}
    \caption{
    \emph{Motivation of AnomalyLMM}. 
    (\emph{left}) Directly matching candidate images with anomalous action is challenging for LMMs. 
    {This difficulty is twofold: visually, the cues are often subtle and local; textually, the critical descriptors usually constitute only a small semantic fraction of a lengthy, complex query.}
    (\emph{right}) Cloze completion is an effective adaptation strategy for LMMs, allowing LMMs to focus on relevant objects and establish fine-grained cross-modal alignment between text and images. 
    \textcolor{nvidiagreen}{Green completions} {in the upper-right box} indicate that completions from LMM \textcolor{nvidiagreen}{align} with the anomalous description, while \textcolor{red}{{red} completions} denote \textcolor{red}{unaligned}.
    }
    \label{fig:motivation}
\end{figure*}
This task poses two key challenges:
\textbf{(1) Fine-grained cross-modal alignment:} Anomalies are often described with nuanced semantics (\eg, ``staggering while clutching their chest''), requiring precise localization of transient visual cues. Existing methods~\cite{yang2024beyond,zeng2024x2vlm} struggle with such alignment, as they primarily focus on coarse-grained attributes (\eg, clothing or pose).
\textbf{(2) Sparse anomaly samples:} Real-world anomaly data are long-tailed and diverse, making it challenging to train or fine-tune models effectively.
While recent Large Multi-modal Models (LMMs)~\cite{li2023blip, alayrac2022flamingo, achiam2023gpt,qwen2.5} show strong general-purpose visual-language reasoning, they are primarily designed for generative tasks rather than discriminative retrieval. Consequently, directly employing LMMs in anomaly search has two deficiencies: (1) There exists a notable domain gap between the general pretraining objectives and the specific demands of anomaly retrieval. (2) There remains a lack of efficient adaptation strategies to effectively leverage LMMs for deployment in few-shot or resource-constrained scenarios, limiting their applicability to anomaly search (see~\reffig{fig:motivation}~(\emph{left})).

In this work, we propose AnomalyLMM, the first framework that strategically adapts LMMs for text-based person anomaly retrieval.
Instead of directly relying on LMMs to retrieve images based on the given text query, we propose a coarse-to-fine retrieval pipeline that first narrows the candidate space and then progressively refines the search using LMMs' reasoning abilities (see~\reffig{fig:motivation}~(\emph{right})).
Specifically, we first employ a general text-to-image retrieval model to produce an initial list of candidate images given a natural language anomaly query. 
These general retrieval models, trained for global cross-modal alignment, are efficient at coarse-level retrieval and serve as a strong foundation for subsequent reasoning.

\textbf{To move beyond coarse retrieval and enable more fine-grained alignment, we reformulate anomaly search as a masked cloze-style reasoning task.} Key verbs and color attributes in the original query are replaced with structured placeholders (\texttt{<~VERB~>}, \texttt{<~COLOR~>}) using a Large Language Model (LLM). This yields a masked query that disentangles subtle behavioral semantics and visual appearance.
Each candidate image is paired with a masked query and fed into an LMM for cloze completion, thereby prompting the model to focus on subtle behavioral, visual appearance, and other anomaly-related cues.
This cloze-style completion transforms each candidate image into a query-specific textual description.
The intuition is that a correctly paired image inherently contains the visual cues (\eg, action verbs or appearance attributes) required to fill the masked placeholders, whereas irrelevant candidates lack such cues and therefore are more likely to produce incorrect completions or generic tokens such as `UNKNOWN'.
To avoid hallucinations, we constrain the LMM to output \texttt{UNKNOWN} when uncertain, ensuring that completions are grounded in the image.
Next, the original query is compared with all image-specific completions using another Large Language Model (LLM). This comparison is performed semantically, considering language variability. Verbs with similar meanings (\eg, ``balancing'' and ``stumbling'') or colors with close hues (\eg, ``dark'' and ``gray'') are treated as partial matches or matches. 
Based on the matching judgments of the same slots, we derive a semantic re-ranking list of candidate images. 
Finally, we jointly take this re-ranking list and the initial ranking list into consideration, yielding the final ranking list. 
Together, these steps form a unified pipeline that reformulates the anomaly retrieval task as a masked completion and re-ranking process. AnomalyLMM enables LMMs to operate in a reasoning-centric fashion without requiring additional training or fine-tuning. Our contributions are as follows:
\begin{itemize}[leftmargin=*]
    \item \textbf{A training-free, coarse-to-fine framework, AnomalyLMM, that adapts LMMs for anomaly search.} In particular, we successfully repurpose the generative knowledge of pre-trained LMMs for a discriminative retrieval task without any fine-tuning. Our framework achieves this through a synergistic three-step pipeline: \textbf{Cloze Generation}, \textbf{Cloze Completion}, and \textbf{Comparison \& Re-ranking}, which associates the discriminative textual descriptors with the local visual cues.

    \item \textbf{Improved performance and verifiable interpretability on the PAB and UCC benchmarks.} Our method not only outperforms existing methods by $1.93$\% and $1.79$\% Recall@1 accuracy, but also provides crucial, interpretable alignments between textual queries and visual behavior, validating its effectiveness and potential for practical deployment.
\end{itemize}


\section{Related Work}\label{sec:related}
\noindent\textbf{Anomaly Search in Public Safety.} Traditional anomaly search approaches rely on handcrafted features~\cite{wang2010anomaly} or reconstruction-based paradigms~\cite{hasan2016learning}, identifying anomalies as statistical deviations from learned normal patterns. Recent advances adopt deep learning for spatio-temporal modeling, including graph networks~\cite{scarselli2008graph} for crowd dynamics and self-supervised learning~\cite{gui2024survey} to exploit unlabeled behavioral data. While methods~\cite{sultani2018real} achieve strong video-level anomaly classification, they operate in a weakly supervised manner, lacking explicit mechanisms to align visual anomalies with textual queries. Following the general person re-identification~\cite{zheng2017discriminatively} and text-guided person search~\cite{zheng2020dual,bai2023rasa}, text-anchored approaches~\cite{shen2023semantics,wang2025cross} attempt cross-modal retrieval but face two limitations: (1) coarse-grained semantics, such as attributes (clothing or pose) or the binary tags (normal or abnormal). Such an annotation limitation overlooks subtle behavioral cues (\eg, \textit{staggering} vs. \textit{falling}), and (2) bias toward normal behaviors due to long-tail distributions. Such models learn more normal behaviors and are prone to predict normal actions. To address the data scarcity, Yang~\etal~introduce the PAB dataset that pioneers text-based person anomaly search to cover more anomaly action types via natural language description~\cite{yang2024beyond}, yet state-of-the-art solutions~\cite{zeng2024x2vlm,yang2024beyond} rely on supervised fine-tuning, which struggles with rare or unseen anomalies. 
Therefore, it remains challenging to design a scalable system for addressing unseen anomaly actions and giving quick responses, which is key to smart city management. In an attempt to generalize anomaly detection systems, we address these gaps by unifying generative knowledge from LMMs with discriminative retrieval.

\noindent\textbf{Fine-grained Cross-modal Retrieval.} This task has gained increasing attention due to its ability to align and retrieve highly specific content across different modalities (\eg, images and text). Early works in cross-modal retrieval primarily focused on coarse-grained alignment, learning shared representations using methods like canonical correlation analysis~\cite{andrew2013deep} and deep metric learning~\cite{lu2017deep}. However, these approaches often struggle with fine-grained distinctions due to their limited ability to capture subtle inter-modal correspondences. Recent advances leverage fine-grained feature interactions to improve retrieval precision. For instance, Faghri~\etal~introduce a hierarchical alignment framework that matches local regions in images with phrases in text~\cite{Fartash2018vse}. Ji~\etal~employ cross-modal attention to dynamically focus on discriminative sub-elements~\cite{zhong2021step}. Transformer-based models~\cite{gorti2022x,zhong2025zero} further enhance fine-grained retrieval by modeling long-range dependencies across modalities. Another line of research explores contrastive learning for fine-grained cross-modal alignment. Methods like~\cite{radford2021learning,chen2024scene,ma2024ea,xing2025context} employ instance- and patch-level contrastive losses to enforce both global and local consistency. 
There are many methods that introduce generative modeling to retrieval~\cite{zhang2024irgen,qu2024tiger,jin2023diffusionret}.
Despite progress, challenges remain in handling highly nuanced queries and mitigating modality-specific biases. Our work addresses these limitations by introducing a training-free cross-modal re-ranking mechanism that enhances fine-grained discriminability while maintaining computational efficiency.

\noindent\textbf{Large Multi-modal Models for Discriminative Tasks.} Recent LMMs~\cite{chen2024internvl,yao2024minicpm,chen2022re} unify vision-language understanding through large-scale pre-training on aligned image-text pairs. 
Some works also explore audio~\cite{deshmukh2023pengi}, video~\cite{maaz2023video,chen2024videollm} and multiple data formats~\cite{wu2024next,ahmat_mvlp_www25}.
Typical models like CLIP~\cite{radford2021learning}and BLIP~\cite{li2023blip} based on contrastive learning enable zero-shot cross-modal retrieval, while generative LMMs (\eg, GPT-4V~\cite{hurst2024gpt}, LLaVA~\cite{liu2023llava}) excel in open-world reasoning via instruction tuning. 
However, adapting these models for fine-grained anomaly search poses the following unique challenges. (1) \textit{Semantic Granularity Gap.} Generative LMMs prioritize holistic scene understanding over localized behavioral anomalies~\cite{radford2021learning,luo_imagescope_www25,chen_dual_www26}; 
(2) \textit{Adaptation Inefficiency.} Existing strategies~\cite{zhang2023finetuning,rong2025backdoor} require full fine-tuning or adding parameters for tuning, such as LoRA~\cite{hu2022lora} and WoRA~\cite{sun2025data}, which are still prohibitively expensive given sparse anomaly data and large GPU memory consumption. 
While prompt-based methods~\cite{zhou2022learning,karthik2023vision,meng2024ranked,ge_mcot_www26} improve attribute alignment, they fail to capture dynamic anomaly saliency (\eg, temporal \textit{collision} cues). 
Our framework innovates by (i) repurposing LMMs' world knowledge for anomaly localization without parameter updates, and (ii) introducing a training-free adaptation protocol to bridge generative and discriminative paradigms.

\begin{figure*}[!t]
    \centering
    \includegraphics[width=\textwidth]{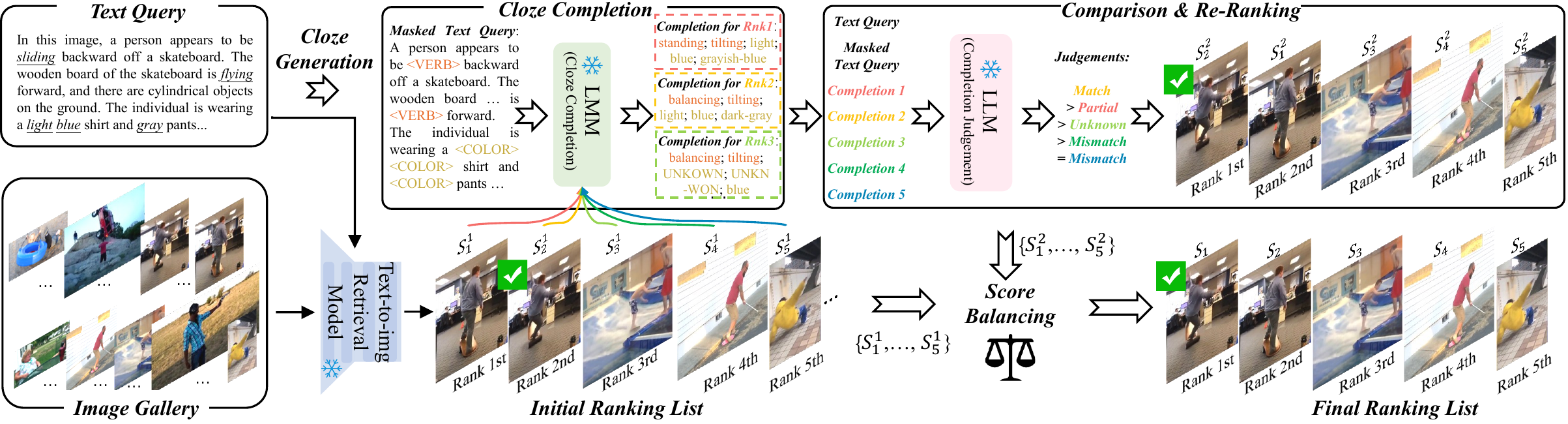}
    \vspace{-.3in}
    \caption{
    \emph{Overall pipeline of the proposed AnomalyLMM.}
    First, given a text query about the target person with anomaly action, we apply an off-the-shelf text-to-image retrieval model to obtain the initial ranking list (\emph{left}).  
    Next, the verbs and colors in the text query are replaced with placeholder tokens $<$~VERB~$>$ and $<$~COLOR~$>$ via an LLM, generating clozes (\emph{Cloze Generation}).  
    According to the candidate images, a pre-trained LMM is then prompted to complete these clozes by predicting masked tokens for each image in the initial ranking list (\emph{Cloze Completion}).  
    These action and color completions represent the LMM's observations of each candidate image for subsequent comparison with the query.
    Subsequently, the completions for each image are compared by an LLM in a semantic equivalent manner, producing the re-ranking list (\emph{Comparison} \& \emph{Re-Ranking}). 
    Finally, the initial and re-ranking lists are combined through score balancing to produce the final ranking list (\emph{upper-right}).
    }
    \label{fig:framework}
    \vspace{-.1in}
\end{figure*}

\section{Method}\label{sec:method}
Our method is a coarse-to-fine retrieval pipeline.
{At the coarse stage,} we first leverage an off-the-shelf text-to-image retrieval model to produce an initial ranking of top-$N$ candidate frames.
{The fine stage consists of} three key steps: Cloze Generation, Cloze Completion, and Comparison \& Re-ranking.
The final ranking result is obtained by fusing the initial ranking list (from the coarse stage) with the re-ranking result (from the fine stage). A brief overview of the pipeline is shown in~\reffig{fig:framework}. We are inspired by the cloze task to bridge the gap between the large models pretrained on generative tasks and the discriminative retrieval demands.  


\subsection{Initial Ranking Generation}
The coarse-to-fine strategy starts with an initial filtering step. Specifically, we employ a dedicated text-to-image retrieval model. This model is trained to match visual content with natural language descriptions and has been widely used in general retrieval tasks due to its efficiency and discriminative power. Given a natural language anomaly query, we leverage the off-the-shelf model to retrieve the top-$N$ images from the dataset according to the retrieval scores $S_n^{1}$. These candidates form an initial ranking list that is expected to include relevant results and exclude most irrelevant ones. 
By reducing the input space for subsequent processing, we not only improve computational efficiency but also allow the LMMs to operate in a more focused, context-rich setting. 
While LMMs excel at contextual reasoning and text-image generation, their native formulation is not directly optimized for discriminative retrieval tasks. 
Thus, instead of directly aligning the text query with visual content using LMMs, we first convert the images in the initial ranking list into text descriptions via cloze generation and completion. This allows us to perform the comparison entirely in the textual domain by matching the generated descriptions with the original query.

\begin{figure*}[!t]
    \centering
    \includegraphics[width=\textwidth]{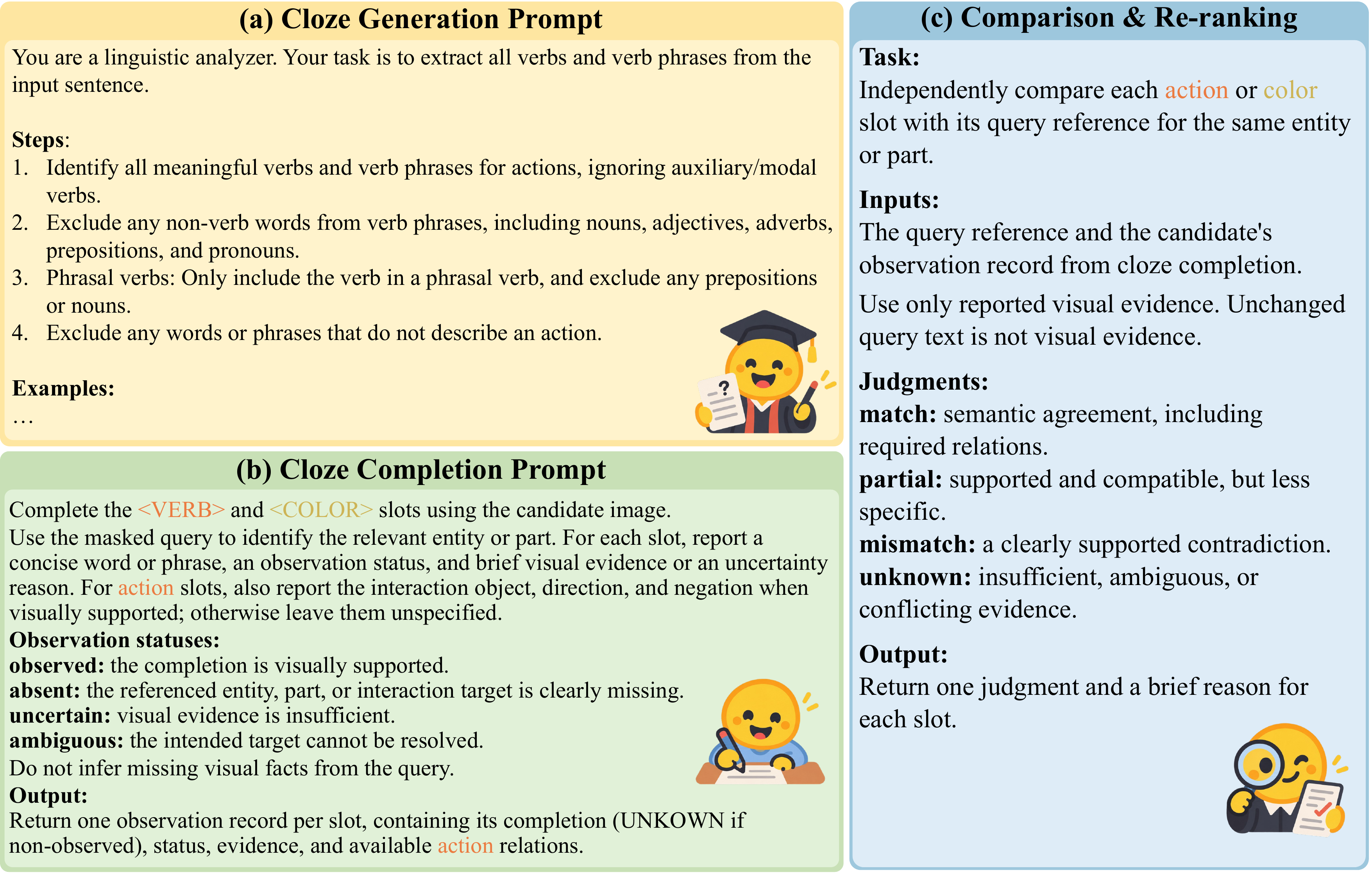}
    \vspace{-.25in}
    \caption{\emph{Prompt for different steps in AnomalyLMM.}
    (a) Cloze generation prompt. This prompt is designed for the LLM to extract verbs that describe actions from input sentences. A similar prompt is used for identifying color-related descriptions.  
    (b) Cloze completion prompt. This prompt guides the LMM to use visual evidence from images to fill the $<$VERB$>$ and $<$COLOR$>$ tokens in the masked text query with contextually appropriate words or ``UNKNOWN'' when no visual support exists.  
    (c) Comparison \& Re-ranking prompt. The LLM independently judges each action or color slot as \emph{match}, \emph{partial}, \emph{mismatch}, or \emph{unknown}. The resulting judgments are converted to scores for re-ranking.
    }
    \label{fig:prompt}
\end{figure*}

\subsection{Cloze Generation}
We leverage LLMs to generate two distinct types of masks from the input text query: verb masks and color masks. These masks are designed to disentangle action semantics and appearance attributes, enabling the model to focus more effectively on salient cues relevant to anomaly retrieval. For the verb mask, we utilize an LLM to identify all action-related or lexical verbs within the query (more can be seen in the supplementary). Each detected verb is then replaced with a standardized placeholder token, \texttt{<~VERB~>}. For the color mask, we prompt the same LLM to extract all color-related descriptions that refer to either the person or their surrounding environment. Identified color terms are similarly substituted with \texttt{<~COLOR~>} tokens one by one. The specific prompts used for action-related verbs are illustrated in~\reffig{fig:prompt}~(a) (the prompt for color-related descriptions is also similar). By combining these two types of masks, we obtain a masked text query, which serves as part of the input to the LMM in the subsequent anomaly-aware retrieval.



\subsection{Cloze Completion}
Given a masked text query containing structured placeholders, \ie, $<\mathrm{VERB}>$ and $<\mathrm{COLOR}>$, we employ an LMM to perform cloze-style completion for each image in the initially retrieved ranking list. 
Specifically, for each candidate image in the ranking list, the LMM is prompted with the image and the obtained masked query as context, and tasked with filling in the missing placeholders based on visual evidence. This process is repeated for all $N$ images in the initial ranking list, resulting in $N$ distinct cloze completions tailored to the visual content of each image. To avoid hallucinated completions, we explicitly instruct the LMM to output \texttt{``UNKNOWN''} if it cannot confidently infer the placeholder content from the given image. Meanwhile, this constraint also helps alleviate the tendency to generate context-independent completions solely based on textual priors rather than grounding them in visual evidence. An example prompt used for this cloze completion step is illustrated in~\reffig{fig:prompt}~(b). These completions are then used for fine-grained re-ranking in the next step. 
Here, our primary objective is to achieve \textbf{synonymous} completions rather than exact word-for-word completions.
Given a correctly paired image, an LMM can complete the cloze task with high semantic fidelity by generating the exact word or its synonyms. In contrast, when the image is unpaired, the model generally fails to produce any semantically appropriate alternatives.




\subsection{Comparison \& Re-Ranking}
We refine the initially retrieved candidates by comparing action and color completions with the original query and balancing the resulting semantic scores with the initial retrieval scores. 

\noindent\textbf{Comparison.}
For each candidate, the LLM is prompted to independently compare each completed action or color slot with its counterpart in the original query.
Each slot receives one of four judgments:
\emph{match} for semantic alignment, \emph{partial} for compatible but less specific evidence, \emph{mismatch} for a clear contradiction, and \emph{unknown} for insufficient or ambiguous evidence.

\noindent\textbf{Re-Ranking.}
Let $v_n^m$ denote the judgment for the $m$-th slot of candidate $n$, \ie, the assigned values $(1,\rho,-1,0)$, corresponding to $(\mathrm{match}$, $\mathrm{partial}$, $\mathrm{mismatch}$, $\mathrm{unknown})$, respectively. $\rho\in[0,1]$ controls the credit for partial matches.
We obtain the semantic alignment score by averaging over all scored action and color slots:
\begin{equation}
    S_n^{2}   = \frac{1}{M}\sum_{m=1}^{M}v_n^m,
\end{equation}
where $M$ is the number of scored slots in the query. We note that different query has different number of slots considering the cloze generation. 
All slots contribute equally, and unknown slots remain included in the denominator. We then combine this score with the initial retrieval score:
\begin{equation}
    S_n = \alpha_1 S_n^{1} + \alpha_2 S_n^{2},
    \quad n\in \{0,\ldots,N-1\},
\end{equation}
where $S_n^{1}$ is normalized retrieval score over the top-$N$ candidates for the query, and $\alpha_1,\alpha_2$ balance the two scores. 
The top-$N$ candidates are sorted by $S_n$ in descending order, with ties resolved by their initial retrieval order.
Candidates outside the top-$N$ are then appended in their initial retrieval order. 
Queries without scored slots retain their initial ranking.




\noindent\textbf{{Discussion-1. What are the advantages of cloze generation and completion?}}
Inspired by recent successes in general text-based person retrieval~\cite{niu2025chatreid,tan2024harnessing,he2024instruct} and target region localization in LMMs~\cite{zhang2025mllms,liu2024rar}, we propose cloze generation and completion to better exploit the reasoning capabilities of LMMs and guide their attention toward anomalous regions in the image.
Compared with previous methods, our cloze generation and completion offer two key advantages: 
(i) Both modules are \emph{training-free} and do not require any additional training or fine-tuning, making them efficient and effective for text-based person anomaly search, especially since real-world anomaly data is scarce and insufficient for fully training LMMs;  
(ii) Cloze generation and completion actively prompt LMMs to infer missing information based on both the image content and the surrounding textual context. This naturally encourages the model to attend to visually relevant regions aligned with the anomaly description, enhancing interpretability and retrieval precision. 

\noindent\textbf{Discussion-2. What are the advantages of comparison \& re-ranking?} 
The proposed comparison \& re-ranking method has three advantages.  
(1) Precise alignment between text query and visual content is critical for anomaly search. Instead of fusing cross-modal features, our method compares cloze completions in the textual domain, enabling fine-grained, interpretable alignment that fully leverages the language understanding capabilities of LLMs.
(2) Due to the inherent ambiguity in natural language, identical actions can be described using different words. Instead of surface-level or lexical matching, our module performs comparison in a semantically equivalent manner, making it more robust to diverse phrasings of the same concept.  
(3) The semantic score of each candidate is derived from explicit judgments on its action and color slots. The four categories distinguish semantic agreement, partial agreement, contradiction, and uncertainty, making the score interpretable. Equal-weight averaging summarizes these judgments, while score fusion combines them with the initial retrieval score to determine the final ranking.

\section{Experiment}\label{sec:exp}
\noindent\textbf{Implementation Details.}
We adopt the official CMP~\cite{yang2024beyond} and the widely-used X$^2$VLM~\cite{zeng2024x2vlm} as two text-to-image retrieval models trained on the PAB dataset. 
All models are fine-tuned with a batch size of 22 using AdamW~\cite{ilya2019adamw} and a weight decay of $1 \times 10^{-2}$. The starting learning rate is set to $5 \times 10^{-5}$ and scheduled by a StepLR policy, with a warmup of $5 \times 10^{3}$ iterations and decays applied at $2 \times 10^{4}$ and $3.5 \times 10^{4}$ iterations.
{We employ Qwen3.8-27B for the cloze generation and the comparison and re-ranking steps, and use the same model for cloze completion. We set $\rho$ to $0.5$.}
\emph{In this work, we do not pursue the best text-to-image model in AnomalyLMM for coarse retrieval, but focus on further improvement by leveraging large models to narrow down the final candidate ranges.}

\noindent\textbf{Benchmarks.} PAB~\cite{yang2024beyond} is the first large-scale benchmark for text-based person anomaly search, covering diverse scenarios
paired with their anomalous counterparts 
for the same identity. 
The test set contains 1,978 real-world image-text pairs with manually verified anomalies, providing a rigorous evaluation platform.
UCC~\cite{yang2024beyond} is built on the UCF-crime dataset~\cite{sultani2018real}, and contains 5,320 real-world image-text pairs in the test set.
We adopt standard retrieval metrics.
Recall@K computes the fraction of queries where the true match appears in the top-K results, reflecting real-world usability. We mainly report Recall@1 (R@1) considering the fine-grained retrieval demands. 
Mean Average Precision (mAP) measures overall ranking quality by averaging precision across recall thresholds, emphasizing correct top-rank retrievals.
All ablation studies are conducted on the PAB dataset.

\begin{table}[t]
  \centering
  \caption{
  \emph{Comparison of proposed method with competitive methods on the PAB~\cite{yang2024beyond} dataset. }
  Integrating representative text-to-image retrieval models into the proposed method consistently improves performance. Both CMP and X$^2$VLM benefit from notable gains in Recall@1 (R@1) and mAP when combined with our approach.
  }
  \vspace{-.1in}
  \tabcolsep=.8cm
    \resizebox{.45\textwidth}{!}{\begin{tabular}{l|cc}
    \shline
    Method & R@1   & mAP \\
    \hline
    APTM~\cite{yang2023towards}   & 72.14  & 82.78  \\
    IRRA~\cite{jiang2023cross}   & 76.39  & 86.33  \\
    CLIP~\cite{radford2021learning}   & 77.60  & 87.35  \\
    RaSa~\cite{bai2023rasa}   & 80.79  & 89.20  \\
    X-VLM~\cite{zeng2021multi}   & 81.95  & 89.86  \\
    \hline
    X$^2$VLM~\cite{zeng2024x2vlm} & 83.77  & 90.43  \\
    X$^2$VLM + \emph{Ours} & \pmb{84.73 (+0.96)}  & \pmb{90.89 (+0.46)}  \\
    {CMP~\cite{yang2024beyond}} & 84.47 & 91.35 \\
    {CMP + \emph{Ours}} & \pmb{86.40 (+1.93)} & \pmb{92.46 (+1.11)} \\
    \shline
    \end{tabular}}%
  \label{tab:sota_pab}%
  \vspace{-.15in}
\end{table}%
\begin{table}[t]
  \centering
  \caption{
  \emph{Comparison of the proposed method with competitive methods on the UCC~\cite{yang2024beyond} dataset.} 
  Integrating the representative text-to-image retrieval model into the proposed method consistently improves performance in Recall@1 (R@1) and mAP.
  }
  \vspace{-.1in}
  \tabcolsep=.8cm
    \resizebox{.45\textwidth}{!}{\begin{tabular}{l|cc}
    \shline
    Method & R@1   & mAP \\
    \hline
    APTM~\cite{yang2023towards}   & 27.86  & 22.61  \\
    IRRA~\cite{jiang2023cross}   & 40.28  & 33.53  \\
    CLIP~\cite{radford2021learning}   & 51.60  & 43.05  \\
    X-VLM~\cite{zeng2021multi}   & 52.33  & 40.87  \\
    RaSa~\cite{bai2023rasa}   & 54.12  & 39.71  \\
    \hline
    CMP~\cite{yang2024beyond}   & 55.31	  & 44.27  \\
    CMP + \emph{Ours} & \pmb{56.35 (+1.04)}    &  \pmb{44.51 (+0.24)} \\
    {X$^2$VLM}~\cite{zeng2024x2vlm} & {62.04}  & {48.25}  \\
    {X$^2$VLM + \emph{Ours}} & \pmb{63.83 (+1.79)}  & \pmb{48.64 (+0.39)}  \\
    \shline
    \end{tabular}}%
  \label{tab:sota_ucc}%
  \vspace{-.2in}
\end{table}%
\begin{figure*}[t]
    \centering
    \includegraphics[width=\textwidth]{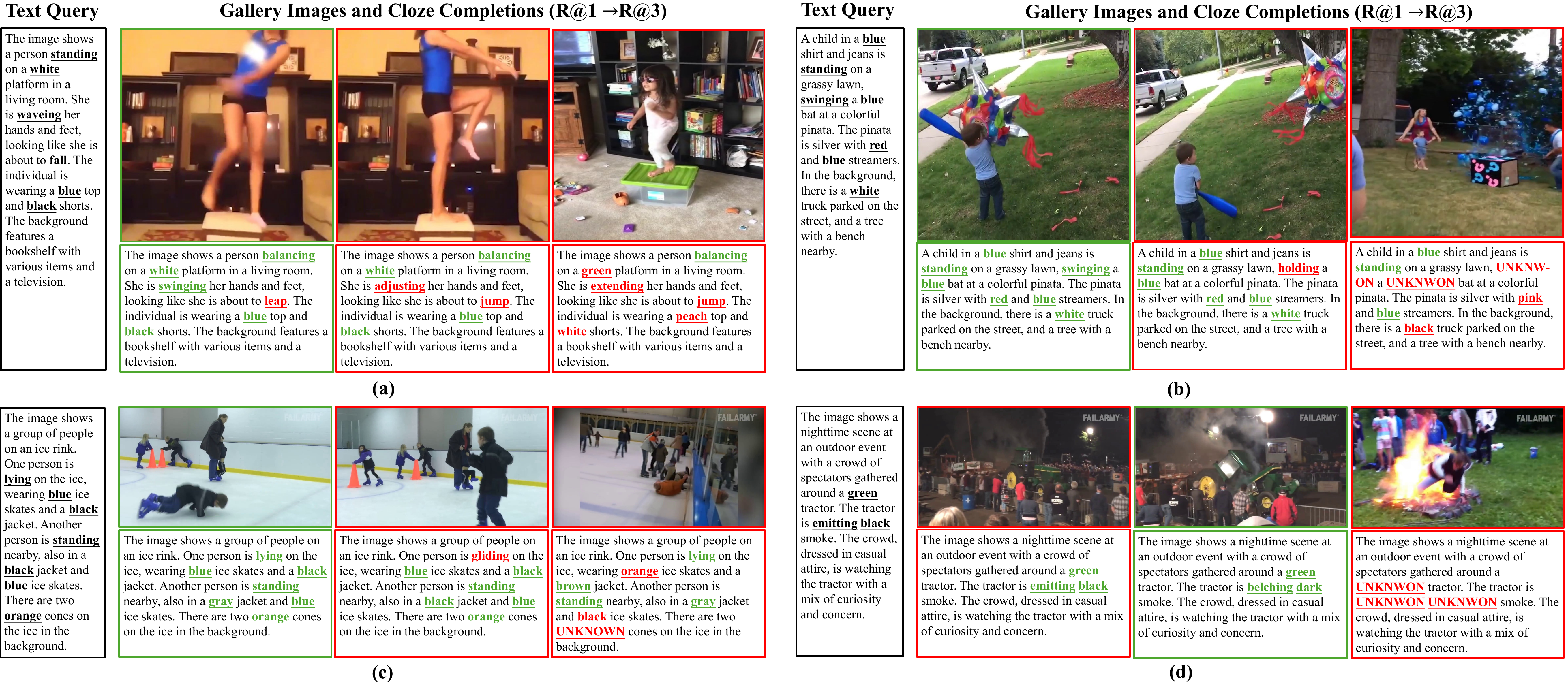}
    \vspace{-.3in}
    \caption{
    \emph{Qualitative results on the PAB dataset.} (a)-(c): successful cases; (d): a failure case.
    Given text queries (left), \textcolor{nvidiagreen}{matched} gallery images and cloze completions are in \textcolor{nvidiagreen}{green}, and \textcolor{red}{mismatched} are in \textcolor{red}{red} (right).
    In (d), the presence of highly similar objects exhibiting comparable appearances and actions consistent with the query text results in a reasonable failure.
    Zoom in for a better visualization.
    }
    \label{fig:visualization}
    \vspace{-.15in}
\end{figure*}

\begin{figure*}[t]
    \centering
    \includegraphics[width=\textwidth]{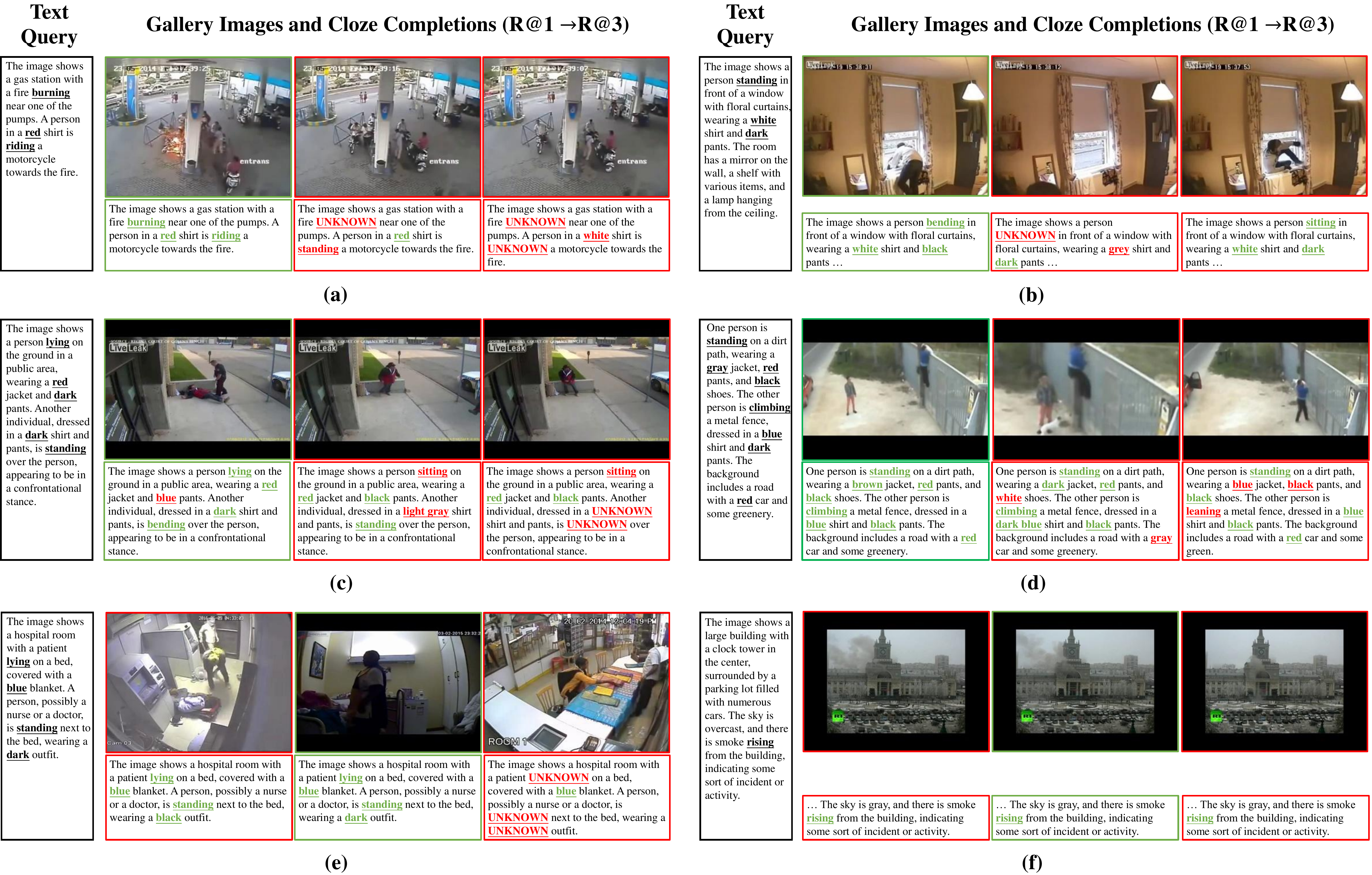}
    \vspace{-.3in}
    \caption{
    \emph{Qualitative results on the UCC dataset.} 
    (a)-(d): successful cases; (e)-(f): failure cases.
    Given text queries (left), \textcolor{nvidiagreen}{matched} gallery images and cloze completions are in \textcolor{nvidiagreen}{green}, and \textcolor{red}{mismatched} are in \textcolor{red}{red} (right).
    In (e), rich textual context provides an overly strong prior, causing the LMM to hallucinate the described scene in unmatched images. 
    In (f), the gallery images are highly similar, containing similar shapes of smoke, resulting in a reasonable failure.
    Zoom in for a better visualization.
    }
    \label{fig:visualization_ucc}
    \vspace{-.1in}
\end{figure*}

\subsection{Comparison with Competitive Methods}
\noindent\textbf{{Quantitative Results}.}
On the PAB dataset, we evaluate the two representative text-to-image retrieval models, \ie, CMP~\cite{yang2024beyond} and X$^2$VLM~\cite{zeng2024x2vlm} in AnomalyLMM, and compare against other competitive approaches (see~\reftab{tab:sota_pab}). 
All competitive methods are fine-tuned on the PAB dataset~\cite{yang2024beyond}.
The proposed method achieves notable improvements over strong baselines~\cite{yang2024beyond,zeng2024x2vlm}. Specifically, we observe $+1.93$\% Recall@1 accuracy, $+1.11$\% mAP on the CMP baseline, and $+0.96$\% Recall@1 accuracy, $+0.46$\% mAP on the X$^2$VLM baseline. 
We observe similar improvement on the UCC dataset (see~\reftab{tab:sota_ucc}). We also adopt CMP and X$^2$VLM as the text-to-image retrieval models in AnomalyLMM. An improvement of $+1.04$\% and $1.79$\% Recall@1 and $+0.24$\% and $0.39$\% mAP is attained on UCC. 
These results validate the effectiveness of cloze generation, completion, and comparison \& re-ranking in the proposed method.

\noindent\textbf{Qualitative Results.}
We present qualitative examples of text-based person anomaly search on the PAB and UCC datasets (see~\reffig{fig:visualization} and~\reffig{fig:visualization_ucc}).  
For cases (a-c) of the PAB dataset in~\reffig{fig:visualization}, our method accurately retrieves the corresponding gallery images based on the query texts. 
In~\reffig{fig:visualization}~(a), actions such as ``waveing'' and ``fall'' are critical.  
The cloze completion for the top-ranked gallery replaces ``waveing'' with ``swinging'', which is semantically similar, while other completions like ``adjusting'' are reasonable but less semantically aligned. However, for the action ``fall'', our method fails to predict accurate completions across all retrieved galleries because the correctly paired image only shows the tendency of an imminent fall. This falling process is visually similar to a leap.
Similar patterns of partial semantic matches and occasional failures appear in~\reffig{fig:visualization}~(b) and~\reffig{fig:visualization}~(c).  
In~\reffig{fig:visualization}~(d), our method does not retrieve the correct gallery image despite semantically matching completions in the top results. This failure is mainly due to the ambiguity in the query text, which presents a challenging scenario even for human annotators to identify the ground truth. 
For cases (a-d) of the UCC dataset in~\reffig{fig:visualization_ucc}, our method successfully retrieves the matched gallery images. 
In~\reffig{fig:visualization_ucc}~(a), the cloze completion keys are the verbs describing the fire and the person near the motorcycle. 
The visual context guides the LMM to successfully predict ``burning'' and ``riding'' for matched images. 
Notably, for unmatched images, the model avoids hallucinations and outputs ``UNKNOWN'' to describe the non-existent fire.
In~\reffig{fig:visualization_ucc}~(b)-(d), the cloze completions target human actions. 
The LMM generates distinct and precise action descriptions, showing that our cloze completion design effectively leverages the world knowledge within LMMs and predicts the action at different granularities.
Failure cases are illustrated in~\reffig{fig:visualization_ucc}~(e)-(f). 
In~\reffig{fig:visualization_ucc}~(e), excessive contextual bias leads to LMM hallucinations, where a person lying near an ATM is misidentified as a patient. This misidentification results in inaccurate completions and suboptimal retrieval. 
In~\reffig{fig:visualization_ucc}~(f), highly similar images in the gallery cause the LMM to generate identical completions, making it difficult to distinguish the matched image in the gallery.

\subsection{Ablation Studies and Further Discussion}

\begin{table*}[!h]
  \centering
  \renewcommand{\arraystretch}{1.1}
  \setlength{\tabcolsep}{4pt}
  \footnotesize
  \caption{
    \emph{Ablation studies on the PAB dataset.}
    (a) Ablation study of primary components. Applying both yields the best performance.
    (b) Ablation study of the score assigned to partial matches. Setting $\rho$=0.5 achieves the best overall performance.
    (c) Ablation study of LLM or LMM for different steps.
    (d) Ablation study on the number of initial candidates.
    (e, f) Ablation study of hyper-parameters ($\alpha_1$, $\alpha_2$) for score balancing.
  }
  \label{tab:all_ablation}
  \vspace{-3ex}

  \begin{tabular}{@{}c@{}}
    \begin{minipage}[t]{0.55\textwidth}
      \centering
      \textbf{(a)}\label{tab:components}\\[3pt]
      \resizebox{\linewidth}{!}{%
        \begin{tabular}{c|c|c|cc}
          \shline
          \multirow{2}[2]{*}{\raisebox{1.5ex}{Method}} & Cloze Gen. \& & Comparison \& & \multirow{2}[2]{*}{\raisebox{1.5ex}{R@1}} & \multirow{2}[2]{*}{\raisebox{1.5ex}{mAP}} \\
          & Completion & Re-Ranking & & \\
          \hline
          Baseline & \textcolor{red}{\ding{55}} & \textcolor{red}{\ding{55}} & 84.47 & 91.35 \\
          $w$ Cloze Completion & \textcolor{nvidiagreen}{\ding{51}} & \textcolor{red}{\ding{55}} & 85.59 & 92.08 \\
          $w$ Re-Ranking & \textcolor{red}{\ding{55}} & \textcolor{nvidiagreen}{\ding{51}} & 84.88 & 91.60 \\
          Ours & \textcolor{nvidiagreen}{\ding{51}} & \textcolor{nvidiagreen}{\ding{51}} & \pmb{86.40} & \pmb{92.46} \\
          \shline
        \end{tabular}%
      }
    \end{minipage}%
    \hspace{0.04\textwidth}%
    \begin{minipage}[t]{0.40\textwidth}
      \centering
      \textbf{(b)}\label{tab:weight}\\[3pt]
      {\normalsize
        \begin{tabularx}{0.90\linewidth}{@{}>{\centering\arraybackslash}X|>{\centering\arraybackslash}p{0.19\linewidth}>{\centering\arraybackslash}p{0.19\linewidth}@{}}
          \shline
          $\rho$ & R@1 & mAP \\
          \hline
          0.70 & 86.34 & 92.44 \\
          0.50 & \pmb{86.40} & \pmb{92.46} \\
          0.30 & \pmb{86.40} & 92.45 \\
          0.10 & 86.34 & 92.43 \\
          \shline
        \end{tabularx}
      }
    \end{minipage} \\[2ex]

    \begin{minipage}[t]{0.55\textwidth}
      \centering
      \textbf{(c)}\label{tab:llm_choice}\\[3pt]
      {\normalsize
        \begin{tabularx}{\linewidth}{@{}>{\centering\arraybackslash}p{0.32\linewidth}|>{\centering\arraybackslash}p{0.32\linewidth}|>{\centering\arraybackslash}X>{\centering\arraybackslash}X@{}}
          \shline
          \multicolumn{2}{c|}{LLM/LMM Selection} & \multicolumn{2}{c}{Performance} \\
          \hline
          \multicolumn{1}{c|}{Cloze Comp.} & \multicolumn{1}{c|}{Re-Ranking} & R@1 & mAP \\
          \hline
          Qwen3.6-35B & Qwen3.6-35B & 86.29 & 92.34 \\
          Qwen3.6-35B & Qwen3.8-27B & 86.24 & 92.31 \\
          Qwen3.8-27B & Qwen3.6-35B & 86.09 & 92.26 \\
          Qwen3.8-27B & Qwen3.8-27B & \pmb{86.40} & \pmb{92.46} \\
          \shline
        \end{tabularx}
      }
    \end{minipage}%
    \hspace{0.04\textwidth}%
    \begin{minipage}[t]{0.40\textwidth}
      \centering
      \textbf{(d)}\label{tab:number}\\[3pt]
      {\normalsize
        \begin{tabularx}{\linewidth}{@{}>{\centering\arraybackslash}X|>{\centering\arraybackslash}p{0.19\linewidth}>{\centering\arraybackslash}p{0.19\linewidth}@{}}
          \shline
          \# Initial Candidate & R@1 & mAP \\
          \hline
          2 & 84.47 & 91.35 \\
          3 & 84.47 & 91.35 \\
          4 & 85.94 & 92.20 \\
          5 & \pmb{86.40} & \pmb{92.46} \\
          \shline
        \end{tabularx}
      }
    \end{minipage} \\[2ex]

    \begin{minipage}[t]{0.47\textwidth}
      \centering
      \textbf{(e)}\label{fig:hyper_parameter_1}\\[3pt]
      \includegraphics[width=\linewidth]{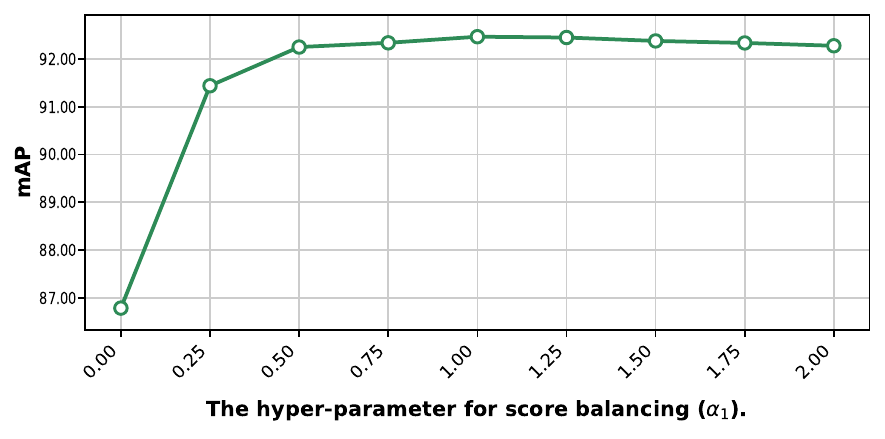}
    \end{minipage}%
    \hspace{0.06\textwidth}%
    \begin{minipage}[t]{0.47\textwidth}
      \centering
      \textbf{(f)}\label{fig:hyper_parameter_2}\\[3pt]
      \includegraphics[width=\linewidth]{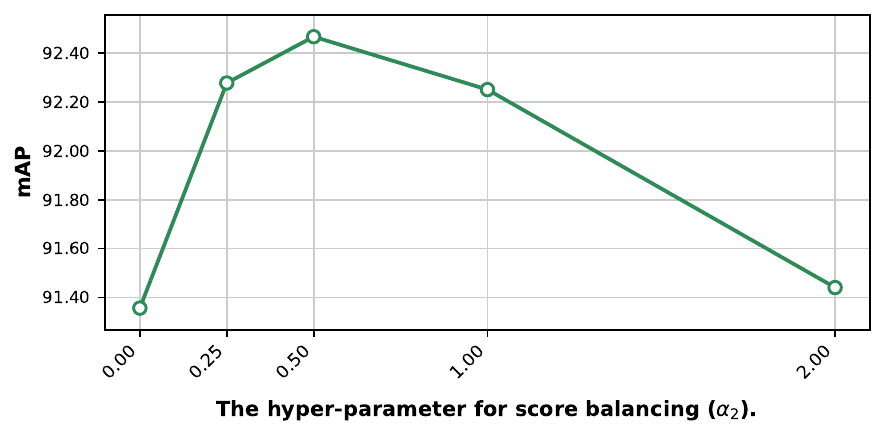}
    \end{minipage}
  \end{tabular}
  \vspace{-4ex}
\end{table*}

\noindent{\textbf{Effect of primary components.}} {We conduct ablation studies with CMP {as the baseline} on the PAB dataset to study our primary components~(see~\reftab{tab:components}~(a)).}
We treat cloze generation and completion as a single module.
{For the effect of {cloze generation and completion} ($w$ cloze completion in~\reftab{tab:components}~(a)), we add this module to the baseline and re-rank generated completions based on the cosine similarity between \texttt{all-MiniLM-L6-v2}~\cite{galli2024performance} embeddings of the completions and the text query.}
{For the effect of the {re-ranking module} ($w$ re-ranking in~\reftab{tab:components}~(a)), we re-rank the initial image list using captions generated by an LMM. The LMM is prompted to focus on object actions and colors.}
With the cloze generation and completion module, $w$ cloze completion attains $+1.12$\% Recall@1 accuracy. With the comparison \& re-ranking module, $w$ re-ranking achieves a $+0.41$\% Recall@1 accuracy. Incorporating both modules, our full method achieves a $+1.93$\% Recall@1 accuracy.

\noindent\textbf{Effect of the score assigned to partial matches.}
We vary $\rho$ from 0.10 to 0.70 while keeping the fusion weights and all other settings fixed (see~\reftab{tab:weight}~(b)).
Setting $\rho=0.50$ achieves the best overall performance.
Although $\rho=0.30$ attains the same R@1, its mAP is slightly lower.
These results support assigning an intermediate score to partial matches.

{\noindent{\textbf{Effect of the choice of LLMs and LMMs in three steps of our method.}}}
We conduct the ablation study on the choice of LLMs in different steps of our method (see~\reftab{tab:llm_choice}~(c)).
Overall, all four configurations achieve comparable performance, with only 0.31\% and 0.20\% variations in R@1 and mAP, respectively, indicating that our method is relatively insensitive to the model choices. 
Notably, using the same model for both steps consistently outperforms mixed-model configurations, suggesting that consistency between cloze completions and subsequent semantic comparisons may benefit retrieval performance.
Among all configurations, Qwen3.8-27B achieves the best results when used in both stages, reaching 86.40\% R@1 and 92.46\% mAP.

\noindent{\textbf{Effect of the number of candidates in the initial ranking list.}}
We conduct an ablation study on the number of candidates in the initial ranking list (see~\reftab{tab:number}~(d)).  
Among the lengths of 2, 3, 4, and 5, re-ranking with a candidate list length of 5 achieves the best performance.  
Using lengths of 2 or 3 leads to a performance drop, while increasing the length to 5 results in a further increase to $86.40$\% Recall@1 (R@1), $92.46$\% mAP. 
These results show the effectiveness of our re-ranking strategy in leveraging more candidates to identify and prioritize the most relevant results.

\noindent{\textbf{Effect of hyper-parameters in score balancing.}}
We investigate the effects of the score-balancing hyper-parameters $\alpha_1$ and $\alpha_2$ (see~\reftab{fig:hyper_parameter_1}~(e) and \reftab{fig:hyper_parameter_2}~(f)).
Overall, combining the initial retrieval and semantic alignment scores achieves the best performance with $\alpha_1=1.0$ and $\alpha_2=0.5$.
Specifically, with $\alpha_2$ fixed at 0.5, increasing $\alpha_1$ from 0 to 1 improves mAP from approximately $86.80\%$ to $92.46\%$, after which the performance remains relatively stable. With $\alpha_1$ fixed at 1, increasing $\alpha_2$ from 0 to 0.5 improves mAP from approximately $91.36\%$ to $92.46\%$, while further increasing $\alpha_2$ to 2 reduces mAP to approximately $91.44\%$.
These findings indicate that the initial retrieval score provides a strong ranking foundation, while the semantic alignment score offers complementary improvements when appropriately weighted.


\section{Conclusion}\label{sec:con}
In this work, we present AnomalyLMM, the first framework to adapt LMMs for text-based person anomaly search. To address the challenges of fine-grained cross-modal alignment and anomaly recognition under sparse samples, our method introduces a novel coarse-to-fine pipeline and training-free adaptation strategies to bridge generative knowledge with discriminative retrieval. In particular, we are inspired by the cloze completion task to harness both LLM and LMM.
Cloze completion makes the observed actions and colors explicit, providing textual evidence for semantic comparison with the query. The resulting slot judgments are averaged with equal weights and fused with the initial retrieval scores to refine the candidate ranking.
Extensive experiments on the PAB and UCC datasets validate significant improvements over competitive baselines, while qualitative analysis validates the interpretability of our anomaly-behavior alignment.

\bibliographystyle{ACM-Reference-Format}
\bibliography{main}







\clearpage
\appendix

\twocolumn[{%
  \begin{center}
    {\LARGE\bfseries
    AnomalyLMM: Bridging Generative Knowledge and Discriminative
    Retrieval for Text-based Person Anomaly Search\par}
    \vspace{0.5em}
    {\Large\scshape Supplementary Material\par}
  \end{center}
  \vspace{0.5em}
}]

\raggedbottom
\section{Appendix}
\noindent
This supplementary material includes {2} aspects:
\begin{enumerate}[nosep,topsep=2pt]
  \item Visualization:
  \begin{itemize}[nosep]
    \item Cloze generation
    \item Cloze completion
  \end{itemize}
  \item Query statistics
  \item Efficiency analysis
\end{enumerate}

\subsection{Visualization}
\label{sec:supp_visualization}
\begin{figure}[H]
    \centering
    \includegraphics[width=\linewidth]{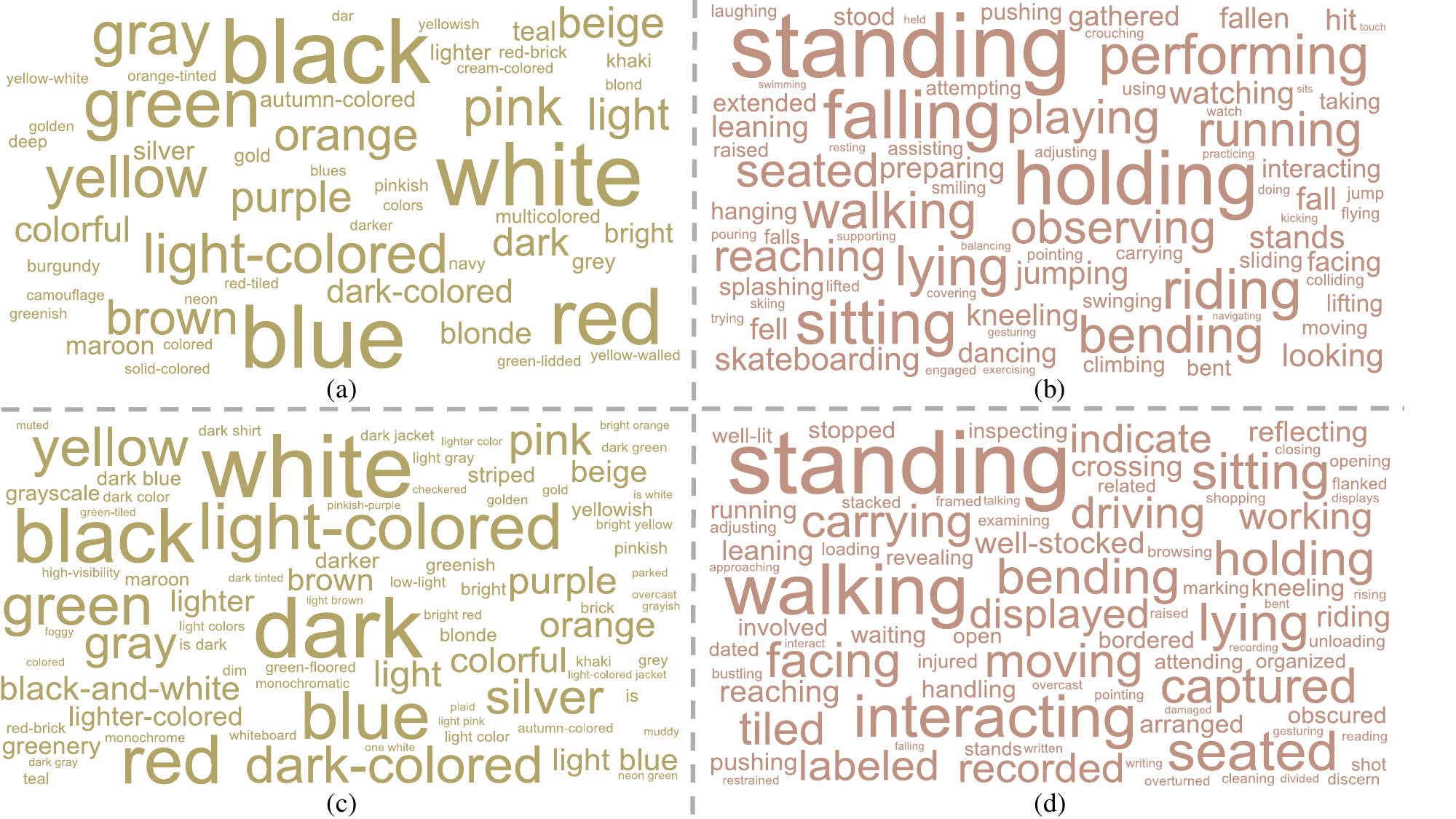}
    \caption{\emph{Word clouds of masked words in PAB~(a,b) and UCC~(c,d).}
    (a,c) Color words. (b,d) Action words.
    }
    \label{fig:supp_mask_word_cloud}
\end{figure}

\begin{figure}[H]
    \centering
    \includegraphics[width=\linewidth]{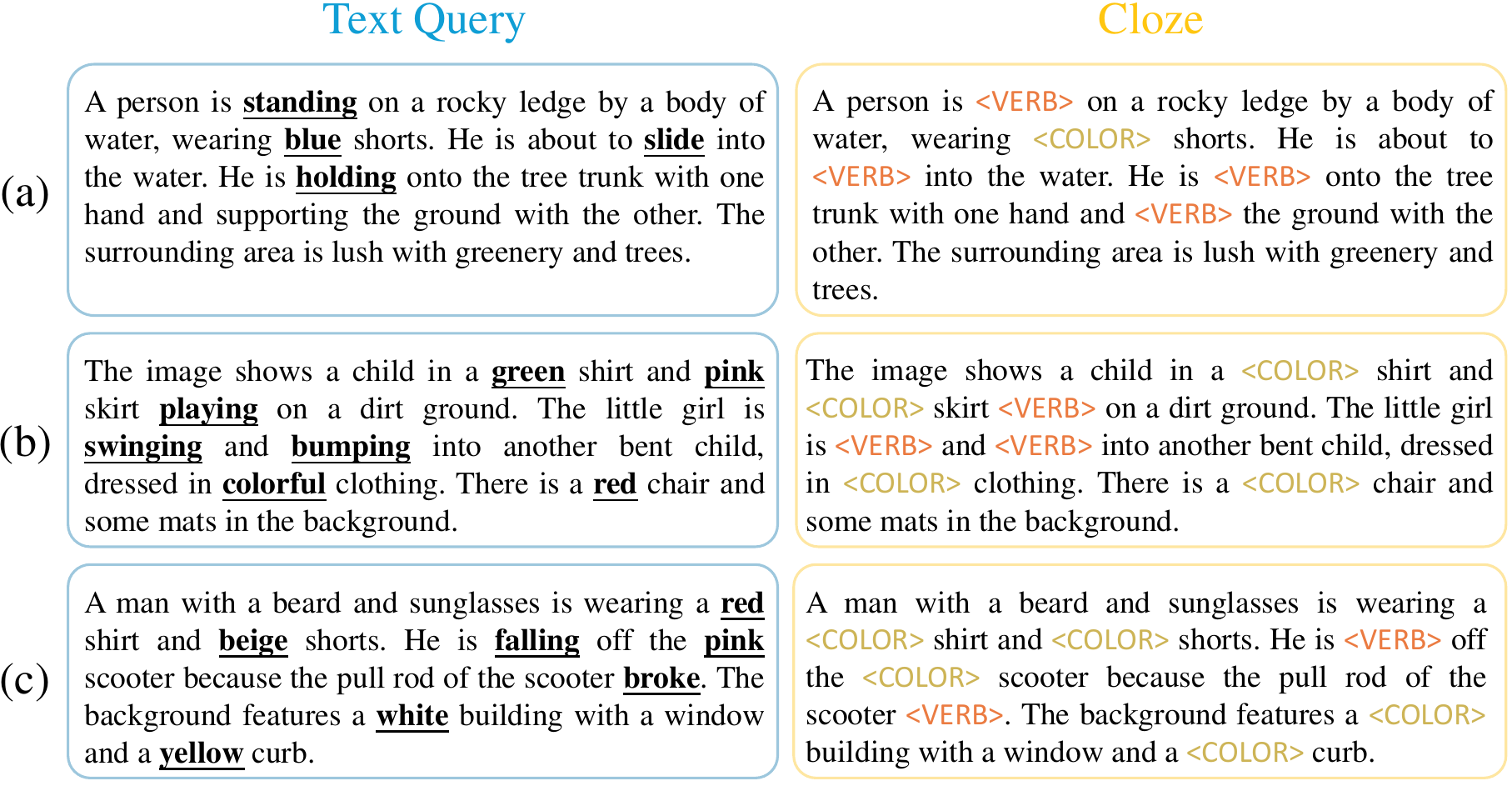}
    \caption{(a--c) \emph{Cloze generation examples.}
    Original queries (left) and clozes (right), with action and color
    words replaced by \textcolor{verbColor}{$<$VERB$>$} and
    \textcolor{colorColor}{$<$COLOR$>$}, respectively.
    }
    \label{fig:supp_cloze_generation}
\end{figure}

\noindent\textbf{Cloze generation.}
We analyze the words selected for action and color slots in PAB, UCC.
\reffig{fig:supp_mask_word_cloud}(a,c) show basic color words, such as black
and white, alongside broader color descriptions, such as light-colored
and dark-colored. Variants such as gray and grey also appear. This
variation in wording and specificity motivates semantic comparison of
color slots.
\reffig{fig:supp_mask_word_cloud}(b,d) includes postures, such as standing,
and anomalous actions, such as falling. Semantic comparison of action slots
must therefore account for the context and action state.
\reftab{tab:supp_dataset} reports averages of 5.94 lexical verbs and
2.52 color words per query in PAB, giving a ratio of approximately
2.36:1. Lexical verbs are more numerous on average, while both types
provide multiple cues within the queries. These statistics count words
in the queries, not masked slots.
\reffig{fig:supp_cloze_generation}~(a), (b), and (c) contain
4, 3, and 2 action slots and 1, 4, and 5 color slots, respectively.
The scope of slots also varies: in (c), action slots refer to the person falling and the scooter's pull
rod breaking, while color slots refer to clothing, the scooter, and
background objects.
The clozes retain the context needed to interpret each slot. Nearby nouns identify the relevant entity or part; ``about to'' in (a) preserves the action state, and ``because'' in (c) preserves the query's causal structure.
This context specifies what to inspect, while visual support must come from the candidate image.
Cloze generation thus exposes multiple action and color slots while
retaining their context. Cloze completion uses each candidate image to
fill these slots, and semantic comparison assesses their agreement with
the original query.

\begin{figure}[H]
    \centering
    \includegraphics[
      width=\linewidth,
      height=0.72\textheight,
      keepaspectratio
    ]{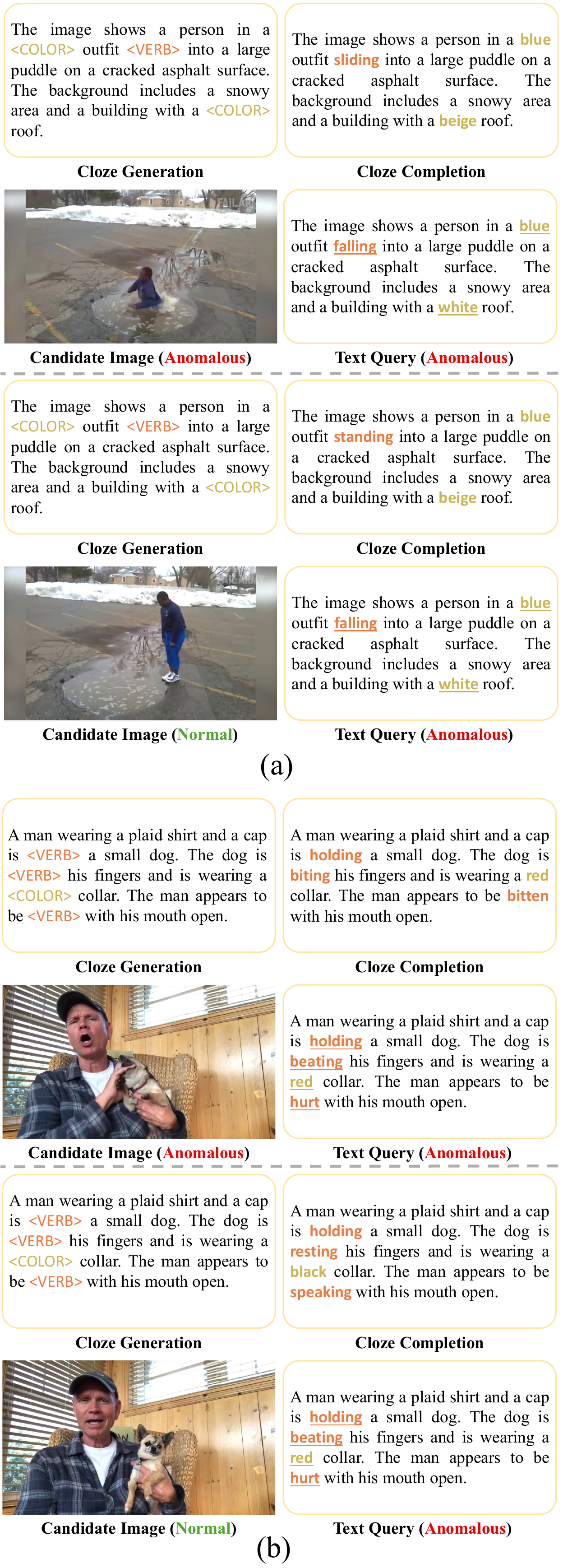}
    \caption{(a,b) \emph{Visualizations of cloze completion}: (upper) Cloze completion for anomalous-action and normal-action candidate.
    }
    \label{fig:supp_cloze_completion}
\end{figure}

\noindent\textbf{Cloze completion.}
In~\reffig{fig:supp_cloze_completion}, we compare completions for paired anomalous and unpaired normal candidates, using the same cloze for each query.
In \reffig{fig:supp_cloze_completion}~(a), both candidates produce ``blue'' for the clothing color slot, but the action slot changes from ``sliding'' to ``standing''. This captures an action difference despite similar clothing and background.
The roof color slot is completed as ``beige'' for both candidates, while the query specifies ``white''. Thus, even a paired candidate can have completions that disagree with the query.
In \reffig{fig:supp_cloze_completion}~(b), both candidates produce ``holding'' for the first action slot. The other action slots change from ``biting'' and ``bitten'' for the paired candidate to ``resting'' and ``speaking'' for the normal candidate.
The color slot also changes from ``red'' to ``black''.
These examples illustrate how cloze completion converts a pre-trained
LMM's generative output into explicit action and color descriptions for
discriminative retrieval. Semantic comparison uses these descriptions
to judge each slot before re-ranking. The completed slots can be checked
against the candidate image and original query, making the alignment
interpretable without fine-tuning the LMM.

\subsection{Query statistics}
\label{sec:supp_query_statics}
\reftab{tab:supp_dataset} reports the average numbers of words,
lexical verbs, and colors per query in PAB and UCC.
Queries in both datasets average more than 50 words and contain multiple lexical verbs and color words. After masking, the remaining query context can help identify the relevant entities and parts, guiding the LMM to complete action and color slots using visual evidence from the candidate
image.
\begin{table}[H]
  \centering
  \caption{
  \emph{Query statistics.} \#words, \#verbs, and \#colors denote the number of words, lexical verbs, and colors per query, respectively.
  }
    \fontsize{8.5}{10}\selectfont 
    \begin{tabular}{l|ccc}
    \shline
    Dataset & \#words & \#verbs & \#colors \\
    \hline
    PAB   & 52.54 & 5.94  & 2.52 \\
    UCC   & 65.43 & 7.16  & 1.45 \\
    \shline
    \end{tabular}
  \label{tab:supp_dataset}%
  \vspace{-5pt}
\end{table}

\subsection{Efficiency analysis}
\label{sec:supp_efficiency}

We report the average processing time per query at each step\reftab{tab:supp_efficency}. Cloze completion is the most time-consuming step and is applied only to the top-$N$ candidates selected by
initial retrieval.
LMMs and LLMs are evaluated on 4 * A100 with VLLM 0.19.0.

\begin{table}[H]
  \centering
  \caption{\emph{Efficiency analysis.}
  Average processing time per query at each step of AnomalyLMM.}
  \label{tab:supp_efficency}

  \fontsize{8.5}{10}\selectfont
  \setlength{\tabcolsep}{8pt}
  \renewcommand{\arraystretch}{1.1}
  \begin{tabular}{@{}lr@{}}
    \toprule
    Step & Time (s/query) \\
    \midrule
    Cloze Generation         & 0.32 \\
    Cloze Completion         & 2.14 \\
    Comparison \& Re-Ranking & 0.22 \\
    \bottomrule
  \end{tabular}
\end{table}

\newpage

\end{document}